\pdfoutput=1

\documentclass[11pt]{article}

\usepackage{EMNLP2023}

\usepackage{times}
\usepackage{latexsym}

\usepackage[T1]{fontenc}

\usepackage[utf8]{inputenc}

\usepackage{microtype}

\usepackage{inconsolata}

\author{First Author \\
  Affiliation / Address line 1 \\
  Affiliation / Address line 2 \\
  Affiliation / Address line 3 \\
  \texttt{email@domain} \\\And
  Second Author \\
  Affiliation / Address line 1 \\
  Affiliation / Address line 2 \\
  Affiliation / Address line 3 \\
  \texttt{email@domain} \\}

\usepackage{microtype}
\usepackage{inconsolata}
\usepackage{url}
\usepackage{linguex}
\usepackage{babel}
\usepackage{natbib}
\usepackage[useregional]{datetime2}
\usepackage{graphicx} 
\usepackage{amsmath}
\usepackage{tcolorbox}
\usepackage{amsfonts}
\usepackage{amssymb}
 \usepackage{mathtools} 
\usepackage{caption}
\usepackage{subcaption}
\usepackage{dsfont}
\usepackage{mathtools}
\usepackage{bbm}
\usepackage{bm}
\usepackage{multirow}
\usepackage{multicol}
\usepackage{natbib}
\usepackage{tikz}
\usepackage{booktabs}
\usepackage{amssymb,amsmath,linguex,amsthm}
\theoremstyle{definition}

\newcommand{\hidden}[1]{}

\title{Learning Semantic Structure through First-Order-Logic Translation}
\author{Akshay Chaturvedi$^{\dagger}$, Nicholas Asher$^{\dagger\ddagger}$\\ 
       $^{\dagger}$IRIT, $^{\ddagger}$CNRS, \\ Toulouse, France }

\date{\today}

\begin{document}
\maketitle

\begin{abstract}
In this paper, we study whether transformer-based language models can extract predicate argument structure from simple sentences. We firstly show that language models sometimes confuse which predicates apply to which objects. To mitigate this, we explore two tasks: question answering (Q/A), and first order logic (FOL) translation, and two regimes, prompting and finetuning. In FOL translation, we finetune several large language models on synthetic datasets designed to gauge their generalization abilities. For Q/A, we finetune encoder models like BERT and RoBERTa and use prompting for LLMs. The results show that FOL translation for LLMs is better suited to learn predicate argument structure.
\end{abstract}

\section{Introduction}\label{sec:intro}

Transformer-based language models (LMs) \cite{vaswani2017attention,bubeck:etal:2023}  have attracted enormous interest because of their language generating capacities and prowess in many NLP tasks. 
We are interested in LMs and their ability to exploit semantic structure for both grasping linguistic meaning and inference. In this paper, we concentrate on an elementary building block of semantic structure, i.e., predicate argument structure.  For example, predicate argument structure determines that the predicate {\em blue} applies to {\em house} and not {\em car} in the sentence: 
\ex. \label{car1} there was a red car in front of a blue house.  

Predicate argument structure also determines which arguments fill which places in two place predicates like {\em in front of}; in our example, it is the car that is in front of the house, not the house in front of the car.


Recovering predicate argument structure is crucial to capturing and reasoning about the meaning of natural language sentences.  If an LM mixes up which object has which property in a premise, it is guaranteed to make errors in reasoning. While semantic structure eventually involves the scopes of operators and quantifiers and verbal modification using tense or adverbial phrases, in this paper, we concentrate simply on capturing the predicate argument structure of basic predicates applied to object denoting nouns like {\em car}.  Our experiments are restricted to simple sentences involving two indefinite noun phrases (NPs) with one or more modifying predicates for each NP. An example of one such sentence is \ref{car1}, containing two noun phrases \emph{a red car} and \emph{a blue house}. The motivation behind using such sentences is to study whether current models are able to capture predicate argument structure in relatively simple scenarios. While being able to capture semantic structure in such cases doesn't necessitate generalization to actual linguistic data, it is an important precursor towards this goal.  Further, synthetic examples also help pinpoint the kinds of difficulties faced by the current models. 

We investigate two approaches for analysing LLMs' ability to capture  predicate argument structure: question answering (Q/A), and translation into a first-order logical (FOL) form. For Q/A, we prompt LLMs to predict a yes/no answer, where for FOL translation we finetune LLM for the task. We also look at the performance of smaller encoder models like BERT~\cite{bert} and RoBERTa~\cite{roberta:2019} using finetuning on Q/A with various predicate argument datasets. The main reason behind also focusing on smaller encoder models is to extend the study of ~\citet{chaturvedi:etal:2022} which looked at these models' performance on a synthetic dataset of $5$ simple templates. Their dataset consists of sentences with two objects having two different colors. This work extends the aforementioned synthetic dataset by incorporating \emph{more properties} for the two objects, \emph{sentence paraphrasing} and \emph{negation}. 

We find that both encoder models and LLMs are able to learn predicate argument structure for simple sentences with just one predicate for each object mentioned.  However, encoder models are unable to generalize to more complex sentences involving more predicates for the two objects mentioned, and even LLMs fail to fully master the predicate argument structure of such sentences.  

For smaller encoder models, finetuning results in a  much higher accuracy when testing on similar patterns to training data but a low accuracy on dissimilar patterns, as a result of overfitting. For LLMs, the FOL translation gives better results in comparison with Q/A prompting, showing that LLMs can generalize from the simple predicate argument structure to more complex sentences. We find that translation approach also has the important advantage of showing when models add hallucinated content, which we argue a Q/A method cannot do. 

In what follows, Section~\ref{sec:motivation} provides motivation and surveys relevant prior work. Section~\ref{sec:preliminaries} gives some preliminaries for our study, while Section~\ref{sec:data} describes our synthetic datasets. Section~\ref{sec:results} gives the results of our experiments. Finally, in Section~\ref{sec: conclusion}, we discuss conclusions and potential future work.




\section{Motivation and Previous Work} \label{sec:motivation}

\citet{chaturvedi:etal:2022} conduct experiments concerning predicate argument structure on encoder models like BERT and RoBERTa \cite{bert,roberta:2019}. As mentioned in Section~\ref{sec:intro}, they construct a dataset of sentences involving two objects with two different colors.  They use 5 different schemas, two of which are reproduced below in \ref{car} and \ref{bat}, where \textit{col1} and \textit{col2} denote two distinct colors.
\ex. 
\a.\label {car} A \textit{col1} car was standing in front of a \textit{col2} house.
\b. \label {bat}  They played with a \textit{col1} ball and \textit{col2} bat.

 Their synthetic dataset contains $1040$ questions (520 ``yes'' and ``no" questions each) on contexts using schemas \ref{car} or \ref{bat} with different color combinations. For each schema, they provide two  questions that are semantically equivalent given the context (e.g. ``Was the car \textit{col1}?" and ``Was there a \textit{col1} car?").  
 They then test whether the models, finetuned on the CoQA dataset \cite{coqa}, could correctly associate the properties with their relevant bearers in a simple question answering task. They find that all the models except RoBERTa-large~\cite{roberta:2019} achieve low accuracy on this simple Q/A test; the encoder models also behave differently and rather strangely with regard to the original and modified questions. We provide their results in Table~\ref{tab:paraphrase} in Appendix for sake of completeness.  We extend this work with in depth analyses and experiments on a range of datasets.

\citet{feng:steinhardt:2023} investigate the binding problem for entity referring expressions (typically proper names) linked by predicates, though they did not investigate the predicate argument binding problem {\em per se} and certainly not in its full generality.  They examine representations in a transformer after attention and linear normalization layers and argue that binding is done though a particular identification vector. In this paper, we look at the binding problem between simple properties conveyed by adjectives and their bearers typically introduced by indefinite noun phrases. Our preliminary experiments indicate that even substantial LLMs do not completely solve this problem.

It is known that LMs can learn features of abstract syntactic representations of natural language sentences like long distance dependencies involving subject verb argument agreement \cite{linzen:etal:2016,goldberg:2019} and the rarer object past participle agreement in languages like French \cite{li:etal:2023}.~\citet{lakretz:etal:2022} show LMs have near perfect performance on short embedded syntactic dependencies but fail on longer distance embedded dependencies. We are interested in whether LMs can learn the mapping from syntax to semantic representations, of which predicate argument structure is the basic building block.   


\citet{dehghani:etal:2018} points out the problems of generalization for transformer models. ~\citet{linc} integrate LLMs with a theorem prover for natural language inference. They first prompt LLMs to translate text to FOL. The resultant FOL is passed to a theorem prover in order to predict an output. In this work, we look at two approaches to extract predicate argument structure from simple sentences: Question-Answering, and FOL translation. For question answering, we look at finetuning for encoder models and prompting for LLMs. Whereas, for FOL translation, we look at finetuning of LLMs.

\section{Preliminaries}  \label{sec:preliminaries}

As mentioned in Section~\ref{sec:motivation}, we are interested to see whether the models can learn, through finetuning or prompting, the structure of a semantic representation.  Given the successes of the models with learning syntactic structure, it is feasible to assume a good level of performance. To this end, we first check whether the results from~\citet{chaturvedi:etal:2022} were just a limitation of smaller encoder models, like BERT and RoBERTa. We ran the same experiment on large language models (LLMs) in the Mistral~\cite{mistral}, Llama-2~\cite{llama2} and Llama-3~\cite{dubey2024llama} families.  
\begin{table}[t]
\centering
\footnotesize
\begin{tabular}{|c|c|c|}
\hline
Model & Org-Acc & Mod-Acc \\
\hline
Mistral-7B & 97.1 (52.1) & 92.5 (42.5)\\
\hline
Llama-2-7B  & 74.2 (55.8) & 80.2 (34.4)\\
\hline
Llama-2-13B &  99.0 (50.1)  & 93.3 (45.6)\\ 
\hline
Llama-3-8B & 81.3 (65.6) & 87.1 (58.3)\\
\hline

\end{tabular}
\caption{Effect of question paraphrasing on the synthetic dataset of \citet{chaturvedi:etal:2022} with different LLMs. Questions of type ``Was the X \textit{col1}?" are referred to as original questions (org) and question of type ``Was there a \textit{col1} X?" is referred to as modified questions (Mod). The number in brackets denote percentage of cases where the model predicted ``no" as the answer.}
\label{tab:paraphrase1}
\end{table}

The results are shown in  Table~\ref{tab:paraphrase1}. Given this table for large language models and Table~\ref{tab:paraphrase} of~\citet{chaturvedi:etal:2022} for smaller encoder models in the Appendix, we hypothesize that model's training and finetuning on a generic question answering task does not force the model to recover the predicate argument structure of the context.

Since models can not master recovering even a simple predicate argument structure,  we need to ask what about a string makes this recovery difficult.  Formal linguistics provides a map $\mu$ from sentences to logical form and predicate argument structure either via an intermediate stage of syntactic structure or directly as in some forms of categorical grammar \cite{steedman:1996}.  To recover predicate argument structure of a sentence, a model will have to learn an algorithm that given an input string has the same output as $\mu$.  By examining $\mu$, we can pinpoint areas of difficulty for learning predicate argument structure. 

 
The predicate argument structures of the sentences used in this work can be expressed with a simple, first order logical formula, consisting of a conjunction of positive atomic formulae of the form $\phi(\alpha_1)$ or $\psi(\alpha_1, \alpha_2)$ where $\alpha_i$ is a constant or a variable representing an individual object.  Proper names introduce constants while indefinite noun phrases like {\em a car} introduce an existentially quantified variable.  For instance, \ref{car1} has the logical form $\exists x \exists y (Red(x) \wedge Car(x) \wedge Blue(y) \wedge House(y) \wedge Infrontof(x,y))$.
 
Given our input strings and target predicate argument structures, there are two kinds of difficulty an algorithm must solve to find the predicate argument structure.  The first has to do with the complexity of a noun phrase (NP) with multiple modifiers.  In languages like English or French, a complex NP like {\em an old, green dirty car}  is syntactically realized in terms of ``modifier depth'' and has the syntactic structure $[old [green [dirty [car]_n]_{np}]_{np}]_{np}]$.  To get the right predicate argument structure with sentences containing such NPs, the model has to ``unwind'' the syntactic tree applying each predicate to the variable that is the argument of the predicate introduced by the noun.  The ability to generalize from simple modification to more complex modification in a general way requires an ability to learn recursion. Without recursion, a model may learn different patterns for combinations of adjectives. However, this approach would require training on most (if not all) possible combinations of adjectives.
 
The other source of difficulty for learning predicate argument structure are long distance dependencies.  Syntactic realizations of these involve for example relative clauses as in:
\ex. \label{long-distance} the car that was next to the green house was red. 

The predicate {\em red} in \ref{long-distance} is not close to its bearer.  Such long distance dependencies have various syntactic representations (e.g. trace binding after movement in theories that allow movement) or require type raising in categorical grammars, which rewrites a function to take an argument of different type \cite{steedman:1996}, to guide $\mu$ to link the predicate with its argument.   

We look at both kinds of difficulty by investigating LM behavior on a variety of synthetic datasets.  These datasets enable us to narrow down what algorithms LMs use to capture predicate argument structure given these two dimensions of complexity. For instance, it is possible for a model to do well say on a Q/A dataset by finding ``short cut'' algorithms instead of the true predicate argument structure.  For instance, the model might guess the right answer to a question like {\em is there a red car?} for a sentence like \ref{car1} simply by computing the shortest distance between a color term and {\em car}, in the context.  But this will fail for examples like \ref{long-distance}.

We use two tests to determine whether the model has correctly grasped the predicate argument structure of the context. The first is a Q/A task that asks about which objects in the context have which properties.  The answers to the questions should completely determine the logical form of the context. It turns out, as we show in Section~\ref{sec:results}, that this is difficult to do.  The second way is for the model to produce the logical form directly. For this task, we give the model a sentence from our synthetic dataset as input and train it to output a correct logical form for that sentence in first order logic.  

A complicating factor in the design of the Q/A experiments is what form the questions in the Q/A task should take.~\citet{chaturvedi:etal:2022} show that encoder models are sensitive to different formulations of a question in Q/A tasks. They behave quite differently when answering questions that are semantically equivalent given the input context, as we can see by comparing Org-Acc and Mod-Acc in Table~\ref{tab:paraphrase}, in the Appendix, for encoder models. We observe the same behavior, althought to a lesser extent, in LLMs as well, as shown in Table~\ref{tab:paraphrase1}.




\section{Datasets and Models} \label{sec:data}

To probe a model's grasp of predicate argument structure in more detail, we develop several synthetic datasets of increasing complexity. The original dataset of \citet{chaturvedi:etal:2022} consists of $5$ templates, each containing two objects of different color. For each template, there were two semantically equivalent questions: original question of the type ``Was the obj col1?'', and modified question of the type ``Was there a col1 obj?'' where col1 and obj refer to color and object respectively. All the $5$ templates have the objects and the corresponding color next to each other (e.g. ``The red car was in front of the blue house''). As a result of this, we observed that when a model is trained on one of the templates, it learns to generalize to other templates as well. To counteract this, we extend the original dataset 
to $25$ templates containing more complex templates with several long distance predicate argument structures as in \ref{long-distance} but also {\em Red was not the color of the car but of the house}. We refer to this dataset as $D_{1,1}$. We list down all the $25$ templates along with their FOL translation in Table~\ref{tab:d1-1} of  Appendix. 
Apart from $D_{1,1}$, we also modify the $5$ original templates to construct more complex datasets, i.e., $D_{2,1}$, $D_{2,2}$, $D_{3,1}$, $D_{3,2}$, $D_{3,3}$; where $D_{i,j}$ refers to the two objects having $i$ and $j$ properties each. We use all these datasets for Q/A and FOL translation task. We do not add templates with complex long distance dependencies to these datasets, as our results indicate that this setting would be too difficult for LLMs to master.  We also construct an additional dataset for question answering, $D_{and}$, where we ask ``and'' questions dataset such as ``Was there a blue car and a red house?'' for the $25$ templates of $D_{1,1}$. All the synthetic datasets have equal number of \emph{yes} and \emph{no} questions.


Along with our synthetic datasets, we also work with the FOLIO dataset~\cite{folio}. This dataset is an NLI dataset but also provides the first-order logic form for premises and hypotheses. We make use of these FOL forms along with a portion of our synthetic dataset to finetune LLMs to generate first order logic from simple sentences. We experiment with four LLMs, namely, Llama-2-7b, Llama-2-13b~\cite{llama2}, Mistral-7b~\cite{mistral}, and Llama-3-8b~\cite{dubey2024llama}. For finetuning these models, we use the QLoRA method~\cite{dettmers2023qlora}. All the models are finetuned for $10$ epochs. We provide details of computing infrastructure and hyperparameters in Table~\ref{tab:model-details} of the Appendix. For question-answering prompting experiments, we give the prompt in the format ``The blue car was standing in front of a red house.$\textbackslash n \textbackslash n$Q: Was there a red car?$\textbackslash n$A:'' where $\textbackslash n$ is the newline character. We observe that adding in-context examples in the prompt does not lead to any improvement in terms of accuracy. This might be because we are prompting LLMs for simple yes/no questions. For encoder models, we use the CoQA finetuned \emph{base} and \emph{large} variants of BERT~\cite{bert} and RoBERTa~\cite{roberta:2019} from~\citet{chaturvedi:etal:2022}. The suite of synthetic datasets along with the best performing LLM finetuned for FOL translation are available here \footnote{\url{https://huggingface.co/akshay107/nl-to-fol}}.

\begin{table}[t]
\centering
\footnotesize
\begin{tabular}{|c|c|c|}
\hline
Model & CoQA & FT \\
\hline\hline
BERT-base    & 56.8 & 99.1  \\
\hline
BERT-large    & 72.2 & 99.8 \\
\hline
RoBERTa-base    & 57.0 & 99.7 \\
\hline
RoBERTa-large    & 84.6 & 99.0 \\ 
\hline
\end{tabular}
\caption{Question-Answer Accuracy for different models on the $D_{1,1}$ dataset. CoQA refers to models finetuned only on CoQA and FT refers to further finetuning on the window/door templates for org question type of $D_{1,1}$. For FT, the score is on the test set of $D_{1,1}$.}
\label{tab:qa-acc-org}
\end{table}

\begin{table}[t]
\centering
\footnotesize
\begin{tabular}{|c|c|c|c|}
\hline
Model & $D_{and}$ & $D_{2,2}$ & $D_{3,3}$  \\
\hline\hline
BERT-base    & 50.1/ 50.0 & 55.2 / 52.3 & 54.2/ 51.3 \\
\hline
BERT-large    & 58.4 / 52.1 & 74.6 / 56.8 &  69.4 /50.0 \\
\hline
RoBERTa-base    & 50.0 / 50.6 & 62.7 / 62.2 & 65.5 / 54.4 \\
\hline
RoBERTa-large    & 63.5 / 50.7 &62.7 / 62.2  & 88.2 / 50.0\\ 
\hline
\end{tabular}
\caption{Question-Answer Accuracy for different models on the $D_{and}$, $D_{2,2}$ and $D_{3,3}$ datasets. The scores in each cell are in the format CoQA/FT.}
\label{tab:qa-acc-complex}
\end{table}

\begin{table*}[h]
\centering
\footnotesize
\begin{tabular}{|c|c|c|c|}
\hline
Model & Dataset & Q/A Accuracy & Pred-Arg Accuracy\\
\hline
\multirow{7}{*}{Mistral-7B}   & $D_{1,1}$ & 87.2 / 94.1 / 80.3 & 59.0 / 78.4 / 39.6 \\
                              & $D_{and}$ & 75.0 / 69.5 / 80.5 & 50.3 / 39.1 / 61.5 \\
                              & $D_{2,1}$ & 86.4 / 89.4 / 83.4 & 41.2 / 45.2 / 37.2 \\
                              & $D_{2,2}$ & 89.5 / 94.0 / 84.9 & 51.7 / 63.5 / 40.0 \\
                              & $D_{3,1}$ & 80.8 / 84.9 / 76.6 & 20.2 / 23.0 / 17.4 \\
                              & $D_{3,2}$ & 86.4 / 92.1 / 80.7 & 31.7 / 41.1 / 22.3 \\
                              & $D_{3,3}$ & 89.3 / 96.1 / 82.4 & 48.5 / 61.8 / 35.3  \\\hline
\multirow{7}{*}{Llama-2-7B}     & $D_{1,1}$ & 69.9 / 74.3 / 65.6 & 25.2 / 32.9 / 17.5  \\
                              & $D_{and}$ & 64.5 / 72.2 / 56.9 & 30.4 / 46.9 / 13.8 \\
                              & $D_{2,1}$ & 69.6 / 74.9 / 64.3 & 11.6 / 14.2 / 8.9 \\
                              & $D_{2,2}$ & 76.4 / 81.2 / 71.7 & 22.7 / 23.1 / 22.3 \\
                              & $D_{3,1}$ & 63.9 / 68.3 / 59.5 & 2.6 / 3.2 / 2.0 \\
                              & $D_{3,2}$ & 69.9 / 75.4 / 64.5 & 6.1 / 8.8 / 3.4 \\
                              & $D_{3,3}$ & 74.1 / 77.6 / 70.6 & 13.4 / 16.9 / 9.8 \\\hline
\multirow{7}{*}{Llama-2-13B}    & $D_{1,1}$ & 89.1 / 91.8 / 86.4 & 62.6 / 72.2 / 53.1 \\
                              & $D_{and}$ & 83.7 / 75.1 / 92.3 & 68.2 / 51.8 / 84.5 \\
                              & $D_{2,1}$ & 90.2 / 93.5 / 86.9 & 51.7 / 61.9 / 41.6 \\
                              & $D_{2,2}$ & 93.2 / 97.5 / 88.9 & 54.6 / 81.5 / 27.7 \\ 
                              & $D_{3,1}$ & 88.1 / 90.5 / 85.7 & 36.1 / 45.2 / 27.0 \\
                              & $D_{3,2}$ & 92.5 / 95.0 / 90.0 & 47.6 / 59.7 / 35.4 \\
                              & $D_{3,3}$ & 93.3 / 94.1 / 92.6 & 49.6 / 56.0 / 43.2 \\\hline
\multirow{7}{*}{Llama-3-8B}    & $D_{1,1}$ & 86.9 / 90.0 / 83.7 & 56.5 / 64.4 / 48.7  \\
                              & $D_{and}$ & 86.1 / 90.5 / 81.7 & 73.9 / 84.0 / 63.8 \\
                              & $D_{2,1}$ & 86.5 / 84.7 / 88.3 & 43.2 / 39.5 / 46.9 \\
                              & $D_{2,2}$ & 84.8 / 81.6 / 88.0 & 35.0 / 27.7 / 42.3 \\ 
                              & $D_{3,1}$ & 84.1 / 84.5 / 83.6 & 23.4 / 26.4 / 20.4 \\
                              & $D_{3,2}$ & 85.1 / 83.8 / 86.4 & 21.8 / 23.8 / 19.7 \\
                              & $D_{3,3}$ & 83.4 / 80.7 / 86.2 & 20.8 / 24.6 / 17.1 \\\hline
\end{tabular}
\caption{Results of Q/A prompting for different LLMs and datasets. The three values in Q/A accuracy and pred-arg accuracy denote the \textbf{overall accuracy/ accuracy for the original question type/ accuracy for the modified question type} respectively. For overall pred-arg accuracy, the model has to predict correctly for all the questions of a particular question type.}
\label{tab:qa-predarg-all}
\end{table*}

\section{Results} \label{sec:results}
\subsection{Results from the Q/A approach} \label{sec:QA}
 
We use our different datasets to  see whether models can generalize from certain kinds of training to new datasets that resembled training data but introduced new elements---more modifications of objects, long distance links, etc.  Generalizability is not guaranteed, as finetuning LLMs can lead to overfitting 
\cite{dehghani:etal:2018}.  


Table~\ref{tab:qa-acc-org} shows the results of encoder models when further finetuned on window/door templates of $D_{1,1}$ for original question type. From the table, we can see that the encoder models, after finetuning, perform better on the augmented synthetic $D_{1,1}$ dataset in comparison to the original models. However, they clearly seem to overfit to the patterns of the $D_{1,1}$, as their performance drops drastically on the $D_{2.2}$ and $D_{3,3}$ datasets as shown in Table \ref{tab:qa-acc-complex}. The question answering task for $D_{2,2}$ and $D_{3,3}$ datasets is more difficult, as the models had to answer $8$ questions for each example for $D_{2,2}$ and $12$ for each example in $D_{3,3}$.  A similar pattern is observed for $D_{add}$ as well in Table \ref{tab:qa-acc-complex}.

Overall, our experiments show that encoder models finetuned on the simple synthetic dataset $D_{1,1}$ fail to generalize to more complex scenarios involving more properties or predicates ascribed to the two objects. Table~\ref{tab:qa-acc-complex} shows that further finetuning leads to near-random accuracy for all the models and they underperform their original counterparts on more complex datasets.

\begin{table*}[h]
\centering
\footnotesize
\begin{tabular}{|c|c|c|c|c|}
\hline
Model & Dataset & FOL Accuracy & Pred-Arg Accuracy & Hallucination Cases\\
\hline
\multirow{6}{*}{Mistral-7B}   & $D_{1,1}$ & 48.0 & 85.7 & 12/440 \\
                              & $D_{2,1}$ & 6.9 & 27.2 & 10/640 \\
                              & $D_{2,2}$ & 0.4 & 22.3 & 0/260 \\
                              & $D_{3,1}$ & 1.8 & 23.0 & 2/1368 \\
                              & $D_{3,2}$ & 3.6 & 21.6 & 4/2088 \\
                              & $D_{3,3}$ & 0.6 & 20.5 & 1/468 \\\hline
\multirow{6}{*}{Llama-2-7B}     & $D_{1,1}$ & 52.5 & 88.6 & 16/440 \\
                              & $D_{2,1}$ & 8.9 & 9.7 & 8/640 \\
                              & $D_{2,2}$ & 11.2 & 11.2 & 8/260 \\
                              & $D_{3,1}$ & 9.7 & 12.2 & 41/1368 \\
                              & $D_{3,2}$ & 5.7 & 7.0 & 57/2088 \\
                              & $D_{3,3}$ & 5.1 & 6.2 & 18/468 \\\hline
\multirow{6}{*}{Llama-2-13B}    & $D_{1,1}$ & 73.4 & 92.3 & 27/440 \\
                              & $D_{2,1}$ & 31.6 & 41.4 & 87/640 \\
                              & $D_{2,2}$ & 33.1 & 44.6 & 42/260 \\
                              & $D_{3,1}$ & 28.1 & 38.3 & 197/1368 \\
                              & $D_{3,2}$ & 25.8 & 36.6 & 358/2088 \\
                              & $D_{3,3}$ & 30.6 & 45.1 & 103/468 \\\hline
\multirow{6}{*}{Llama-3-8B}    & $D_{1,1}$ & 81.4 & 90.5 & 33/440 \\
                              & $D_{2,1}$ & 51.7 & 58.4 & 126/640 \\
                              & $D_{2,2}$ & 60.8 & 70.0 & 53/260 \\
                              & $D_{3,1}$ & 37.4 & 42.0 & 100/1368 \\
                              & $D_{3,2}$ & 38.9 & 41.7 & 234/2088 \\
                              & $D_{3,3}$ & 28.6 & 29.1 & 52/468 \\\hline
\multirow{6}{*}{Llama-3-8B*}    & $D_{1,1}$ & 84.1 & 91.8 & 62/440 \\
                              & $D_{2,1}$ & 64.1 & 67.4 & 135/512 \\
                              & $D_{2,2}$ & 73.1 & 77.7 & 66/260 \\
                              & $D_{3,1}$ & 47.6 & 50.1 & 165/1368 \\
                              & $D_{3,2}$ & 52.1 & 56.1 & 275/2088 \\
                              & $D_{3,3}$ & 50.9 & 51.7 & 69/468 \\\hline
\end{tabular}
\caption{Results of FOL approach for different fine tuned models and datasets. The first four models were trained on FOLIO and window/door templates of $D_{1,1}$. However, llama3-8B* was trained on FOLIO, window/door templates of $D_{1,1}$ and $D_{2,1}$. The scores are provided on the test set. Hallucination cases are shown in the format $\#cases/\#total$.}
\label{tab:qa-fol-all}
\end{table*}


In addition, we note that the scores in Table~\ref{tab:qa-acc-org} give accuracy for all the questions.  This is not sufficient to determine predicate argument structure. In order to do so, we need the model to answer all questions for a particular input correctly. We refer to this accuracy as \emph{pred-arg accuracy}. For pred-arg accuracy, the model needs to answer all $4$ questions for a $D_{1,1}$ example correctly; whereas for a more complex dataset like $D_{3,3}$, it needs to answer all $12$ yes/no questions correctly. Given the prefect scores in Table \ref{tab:qa-acc-org}, the models will have a very high pred-arg accuracy on $D_{1,1}$, but not on more complex datasets, as evident from Table~\ref{tab:qa-acc-complex}.



Table~\ref{tab:qa-predarg-all} shows the result of Q/A prompting for different LLMs. We report both Q/A as well as pred-arg accuracy. For both cases, overall as well as question type specific accuracy is provided. From the table, we can see that prompting leads to relatively good but not perfect scores on $D_{1,1}$. Prompting also provides a more graceful decline in performance over the more complex datasets. However, there are some peculiarities.  Llama-2-7B and Llama-3-8b have low pred-arg accuracy, even though both the models have good overall Q/A accuracy. An exception is $D_{and}$ dataset for LLama-3-8B where the model achieves good overall pred-arg accuracy.  Across the LLMs, Llama2-13b performed the best indicating that a larger number of parameters helps on the more complex datasets, especially for pred-arg accuracy. Llama2-13b's overall pred-arg accuracy on the complex datasets like $D_{3,1}, D_{3,2}, D_{3,3}$ is often more than double that of Llama-3-8B and triple that of Llama 2-7B. Mistral-7B also achieves a decent overall pred-arg accuracy across different datasets.

As mentioned earlier, Table~\ref{tab:qa-predarg-all} also shows the Q/A and pred-arg accuracy separately for each question type across datasets and LLMs. Here, we see that, all the models treat the two semantically equivalent questions quite differently, as there is a significant difference in the two scores.  This raises serious concerns about the robustness and reliability of the Q/A method, as well as theoretical issues about an LLM's grasp of the meaning of questions and question equivalence.


\subsection{Results on the translation task} \label{sec:fol}


\label{tab:fol}

\hidden{\begin{table}[h]
\centering
\footnotesize
\begin{tabular}{|c|c|c|}
\hline
Model & Correct & Incorrect \\
\hline
Mistral-7B & 113 & 31\\
\hline
Llama-2-7B  & 119 & 129 \\
\hline
Llama-2-13B & 119 & 45\\ 
\hline
LLama-3-8B & 182 & 110\\
\hline
\end{tabular}
\caption{Correct/Incorrect for two-place predicates amongst those which were correct as per FOL Accuracy. The cases where a two-place predicate was missing in the predicted FOL also counts towards Incorrect.}
\label{tab:fol-2-pred}
\end{table}}

\hidden{\begin{table}[h]
\centering
\footnotesize
\begin{tabular}{|c|c|}
\hline
Model & Accuracy \\
\hline
Llama7B  & 88.6\% \\
\hline
Llama13B &  92.3\%\\ 
\hline
Mistral-7B & 85.7\%\\
\hline
\end{tabular}
\caption{Accuracy for extracting pred-arg based on predicted FOL {\color {teal} for the D1-1 dataset}. The models were fine tuned on all the window/door templates}
\label{tab:fol-pred-arg}
\end{table}
}

For FOL translation, all the LLMs are finetuned on FOLIO train set and window/door template of $D_{1,1}$. Table \ref{tab:qa-fol-all} shows two accuracy scores for the translation task: an accuracy score calculated relative to an exact match of the FOL translation with a gold standard (i.e., FOL accuracy) and the accuracy of the predicate argument structure that we can algorithmically infer from the LLM's predicted FOL (i.e. Pred-Arg accuracy). The discrepancy between the two scores shows that LLMs stray rather frequently from the translation paradigm for our synthetic dataset.  One main problem problem has to do with an LLM strategy of {\em glueing predicates} together to avoid conjunctions.  For instance {\em big, red, shiny car} might be translated as $BigRedShiny(x) \wedge Car(x)$ or simply as $BigRedShiny(Car)$ instead of the desired $Big(x) \wedge Red(x) \wedge Shiny(x) \wedge Car(x)$. The FOLIO dataset sometimes exemplifies this glueing strategy for complex predicates with modifiers like {\em very large}; so we hypothesize that the LLMs are learning this behavior during finetuning.  Even though these glued predicate translations are not correct first order logic formulas, they can, however, easily be converted to the appropriate logical form.

The lack of accuracy for predicate argument structure based on translation surfaces in three other ways.  First, the LLM's translation may {\em drop} certain predicates from logical form.  For instance {\em big, red, shiny car} might be translated as $Big(x) \wedge Red(x) \wedge Car(x)$ or some other subsequence of the desired translation.  Second, relational predicates like {\em in front of} as in {\em the big car in front of the old house} are not rendered correctly.  A third source of difficulty is that  sometimes models mess up the quantificational structure (though this happens rarely with Llama-2-13b or Llama-3-8b).  

A final and major problem is that hallucinated content is sometimes added to the translation.  In particular, Llama2-13b and Llama3-8b have significant rates of hallucination ($> 10\%$ of the cases on the $D_{3,2}$ dataset).  For example,
\begin{enumerate}
    \item ``A clean red glass was placed on a modern dirty white table." $\exists x \exists y \exists z (Glass(x) \wedge Red(x) \wedge Clean(x) \wedge Table(y) \wedge Modern(y) \wedge White(y) \wedge Dirty(y) \wedge DontMindIfImWhite(z) \wedge Table(z))$
    \item ``A vintage blue glass was placed on a modern dirty red table." $\exists x \exists y \exists z (Vintage(x) \wedge BlueGlass(x) \wedge Table(x) \wedge Modern(x) \wedge Dirty(x) \wedge Red(x) \wedge Vintage(y) \wedge BlueGlass(y) \wedge Cabinet(y) \wedge Queen(y) \wedge Vintage(z) \wedge BlueGlass(z) \wedge Toilet(z) \wedge King(z))$
\end{enumerate}
Table \ref{tab:qa-fol-all} gives figures for hallucination cases as well.

We also note some interesting peculiarities about these failings.  A model may adopt the glueing predicate strategy for instances of one template but not another.  Further, if a model adopts a glueing predicate strategy, normally it hallucinates very little. The difference between the two accuracies (FOL and Pred-Arg) is an indicator of the extent of glueing in model's predicted FOL. As we see from the table, the small models also produce far fewer hallucinations than the larger models; although their accuracy score wae significantly low.  Mistral-7B has very few hallucination cases as it mostly relied on glueing predicates.

In Section~\ref{sec:QA}, we saw that finetuning smaller encoder models for the Q/A task on our synthetic led to overfitting to the training pattern.  Unfortunately, we see a similar pattern especially with the small LLMs on their predictions for the FOL task when we move from the simple dataset to more complex ones. While results are very good on the $D_{1,1}$ dataset, the smaller models fail pretty much completely once we have two modifiers of one noun in the $D_{2,1}$ dataset; they fail to produce anything meaningful on $D_{2,1}$ or more complex datasets $D_{2,2}, D_{3,1}, D_{3,2}, D_{3,3}$. Even Llama-2-13b and Llama-3-8B, perform considerably worse when we move to the more complex datasets .    

Nevertheless, Llama-2-13b and Llama-3-8b show considerable generalization ability on the complex datasets, as they achieve an accuracy score in the double digits. For complex datasets, these models also fail to mention some of the predicates that should be in the logical form.  In these cases, we can't reconstruct the proper logical form, thereby resulting in lower accuracy.  More particularly, Llama-2-13b has a drastic performance drop from $D_{1,1}$ to $D_{2,1}$.  $D_{2,1}$ is the simplest dataset after $D_{1,1}$.  On the other hand, it performs similar on the balanced sets $D_{2,2}$ and $D_{3,3}$.  The imbalanced datasets $D_{3,2}$ and $D_{3,1}$ is more challenging for the model than the balanced dataset $D_{3,3}$. 

Llama-3-8B's performance still attains good accuracy for $D_{2,2}$ but drops significantly for the imbalanced dataset $D_{2,1}$.  Its performance drops further once we move to $D_{3,1}$ and $D_{3,2}$.  Unlike Llama-2-13B, Llama-3-8B has the lowest performance on the $D_{3,3}$ dataset.  In order to address the issue of imbalanced datasets, we augmented the training set for Llama-3-8B by also including window/door templates of $D_{2,1}$.  The resultant model is referred to as Llama3-8b* in Table~\ref{tab:qa-fol-all}. We can see that Llama3-8b* has much higher accuracy on $D_{2,1}$ but, not to the level of $D_{1,1}$.  One possible reason for this is Llama-3's very high ($\sim$ 25\%) hallucination rate on that dataset.  However, its accuracy for the rest of the complex datasets increases considerably, surpassing its counterpart, Llama-3-8B, as well as Llama-2-13B, by a large margin.


\subsection{Comparisons between Q/A and FOL translation}

Both Q/A and FOL translation show that models were able to solve difficulties with predicate argument structures that come from long distance dependencies.  However, with respect to difficulties that stem from depth of embedding, these methods have quite different characteristics.  Finetuning encoder models for the Q/A task provided essentially perfect accuracy on the $D_{1,1}$ dataset.  However, the scores plummeted when the finetuned models had to answer simple yes/no questions on more complex datasets like $D_{2,2}$ or $D_{3,3}$. The scores even for LLMs dropped considerably in terms of pred-arg accuracy, as seen in Table \ref{tab:qa-predarg-all}.  

Furthermore, encoder models and LLM performance on Q/A tasks is quite unstable in that semantically irrelevant differences in the surface form of the question affects how the models respond.  Finally, even an exhaustive Q/A as in Table \ref{tab:qa-predarg-all} only offers a partial view of the predicate argument structure; it tells us what predicates that are mentioned go with which arguments are mentioned, but it does not prevent the model from adding hallucinatory content of an arbitrary nature.  This hallucinatory content can clearly affect downstream reasoning tasks.  Given the random nature of the hallucinatory content, we see no finite way of using Q/A to eliminate the possibility of hallucinated content.  Therefore, we conclude that Q/A is not an optimal way to probe for mastery of predicate argument structure or for models to learn it. The FOL translation task, on the other hand, makes hallucinatory content immediately obvious and also completely determines predicate argument structure.  


\subsection{Comparisons between finetuning and prompting for pred-arg structure}

Prompting for the Q/A strategy with LLMs yields a substantially lower overall pred-arg accuracy in comparison to their FOL finetuned counterparts and even smaller finetuned models like BERT, and RoBERTa.  If we compare predicate-argument accuracy for a given model (even though the two tasks do not give equivalent results because of the presence of unanticipated hallucinatory content), we see that finetuning for translation on Mistral gives much worse results than prompting on Q/A. For Llama-2-7B, the results for the two approaches are equally bad.  For Llama2-13B, prompting achieves higher scores especially for the $D_{2,1}$ and $D_{2,2}$ datasets. However, for these datasets, the scores vary significantly for the two question types.  Llama-3-8b does better with FOL finetuning. Given the increased difficulty of the translation task, we conclude that the finetuning provides better results, in spite of the overfitting problem. From this, we conclude that finetuning at least for predicate argument structure shows evidence of higher generalizability than prompting. Further evidence for our conclusion is the fact that, across the two tasks, Llama-3-8b* has the best pred-arg accuracy score for majority of the datasets.

\hidden{
\begin{table}[h]
\centering
\footnotesize
\begin{tabular}{|c|c|}
\hline
Model & Accuracy \\
\hline
Llama7B  & 69.9\% \\
\hline
Llama13B &  89.1\%\\ 
\hline
Mistral-7B & 87.2\%\\
\hline
\end{tabular}
\caption{Overall Question-Answer Accuracy of models based on \textbf{prompting} on $D_{1,1}$.\textcolor{blue}{Also give accuracy with all 4 correct constraint.}}
\label{tab:qa-prompting}
\end{table}

\begin{table}[h]
\centering
\footnotesize
\begin{tabular}{|c|c|c|c|}
\hline
Model &   $D_{and}$ & $D_{2,2}$ & $D_{3,3}$\\ \hline
\hline
Llama7B  & 64.5\% & 76.4\% & 66.7\%\\
\hline
Llama13B &  83.7\% & 93.2\% & 84.0\%\\ 
\hline
Mistral-7B & 75.0\% & 89.5\% & 80.3\%\\
\hline
\end{tabular}
\caption{Overall Question-Answer Accuracy of models based on \textbf{prompting} on different datasets. {\color {magenta} all 8 all 12 correct score}}
\label{tab:qa-prompting-prop}
\end{table}}

\section{Conclusions} \label{sec: conclusion}
In this paper, we investigate smaller encoder models and  moderate sized LLM's grasp of predicate argument structure for simple sentences.  We use two types of methods for finding predicate argument structure: Q/A and first order logic (FOL) translation and examined their behavior on our synthetic datasets. The results show that neither of the two approaches succeed in mastering this fundamental aspect of meaning. After finetuning, encoder models still could not generalize from their training to find predicate argument structure of more complex sentences.  However, LLMs does manage to show considerable generalization ability with finetuning on the FOL translation task.  Overall, we find that finetuning on the translation task gave the best results for learning predicate argument structure.   

Our results also show that LLMs (especially Llama-2-13B, and Llama-3-8B) tend to hallucinate when finetuned for FOL translation. Recent work has shown that a RAG-based approach can help mitigate hallucination~\cite{ayala-bechard-2024-reducing}. This could be a possible avenue for future work, where a RAG-based approach limits the predicates a model can use in its translation into FOL.

We have discussed prompting vs. finetuning on models of the same size.  However, for much larger models, or models that are not open, finetuning is not an option.  \citet{chaturvedi:etal:2022} show that the GPT-instruct models~\cite{instruct-gpt}, namely, text-davinci-002, and text-davinci-003; achieve near-perfect accuracy on their basic predicate-argument dataset.  We plan to test the latest GPT models on our more complex datasets in the future. 


\section*{Limitations}
In this work, we investigate the models' ability to capture predicate argument structure with embedded modifiers and long distance dependencies. However, we do not put the two difficulties together, say in trying the models out on sentences with two predicates for each argument, at least one of which was connected via a long distance dependency.  As suggested by our results, we suspect that finding the predicate argument structure of such sentences would be very difficult for all the models we tested.  We also do not test how logical operators might affect predicate argument structure as our models already have difficulty just with the simple affirmative contexts.

We find that, for encoder models, the latent representations of relevant predicates have higher weight in the internal representations of their arguments.  We try to reinforce this property with a loss function using a method inspired by \citet{raissi:etal:2017}.  But, this approach doesn't enhance the Q/A accuracy of the model.  Even finetuning on an exogenous source of information from FOL translation does not achieve full mastery of the predicate argument structure of the simple sentences in our synthetic datasets.


\section*{Ethics Statement}
This work shows that LLMs finetuned on FOL translation tend to hallucinate for some cases. The hallucinated content is often completely unrelated to the input text. This poses a serious challenge which needs to be addressed before deploying such models in practice. The lack of robustness of LLMs across semantically equivalent questions is also a detriment to their applicability in the real world.

\section*{Acknowledgement}

For financial support, we thank the National Interdisciplinary Artificial Intelligence Institute ANITI (Artificial and Natural Intelligence Toulouse Institute), funded by the French ‘Investing for the Future– PIA3’ program under the Grant agreement ANR-19-PI3A-000.   This project has been funded by the French government as part of France 2030 and is funded by the European Union - Next Generation EU as part of the France Relance. This research is also supported by the Indo-French Centre for the Promotion of Advanced Research (IFCPAR/CEFIPRA) through Project No. 6702-2 and  Science and Engineering Research Board (SERB), Dept. of Science and Technology (DST), Govt. of India through Grant File No. SPR/2020/000495. This work was granted access to the HPC resources of CALMIP supercomputing center under the allocation 2016-P23060.

\newpage
\onecolumn
\section{Appendix}\label{sec:appendix}

\begin{table*}[h]
\centering
\footnotesize
\begin{tabular}{|c|c|c|}
\hline
Model & Org-Acc & Mod-Acc \\
\hline\hline
BERT-base    & 50.0 (100.0) & 69.4 (59.8)  \\
\hline
BERT-large    & 95.2 (51.0) & 77.3 (27.3)  \\
\hline
RoBERTa-base    & 51.0 (99.0) & 70.0 (78.5) \\
\hline
RoBERTa-large    & 99.4 (49.4) & 95.0 (45.0) \\ 
\hline
XLNet-base    & 50.6 (6.0) & 50.8 (0.7) \\
\hline
XLNet-large    & 75.2 (74.8) & 79.8 (36.3) \\ \hline
\end{tabular}
\caption{Effect of question paraphrasing on different models of \citet{chaturvedi:etal:2022}. Questions of type ``Was the X \textit{col1}?" are referred to as original questions (org) and question of type ``Was there a \textit{col1} X?" is referred to as modified questions (Mod). The number in brackets denote percentage of cases where the model predicted ``no" as the answer.}
\label{tab:paraphrase}
\end{table*}

\begin{table*}[h]
\centering
\tiny	
\begin{tabular}{|c|c|}
\hline
Template & FOL \\ \hline
The col1 car was standing in front of a col2 house. & $\exists x \exists y $(Car(x) $\wedge$ col1(x) $\wedge$ House(y) $\wedge$ col2(y) $\wedge$ standing-in-front-of(x,y))\\
The car that was col1 was standing in front of a house that was col2. & $\exists x \exists y $(Car(x) $\wedge$ col1(x) $\wedge$ House(y) $\wedge$ col2(y) $\wedge$ standing-in-front-of(x,y))\\
col2 was not the color of the car but of the house. & $\exists x \exists y $(Car(x) $\wedge$ $\neg$ col2(x) $\wedge$ House(y) $\wedge$ col2(y))\\
col1 was the color of the car in front of col2 house.& $\exists x \exists y $(Car(x) $\wedge$ col1(x) $\wedge$ House(y) $\wedge$ col2(y) $\wedge$ infrontof(x,y))\\
The car that was in front of the col2 house was col1.& $\exists x \exists y $(Car(x) $\wedge$ col1(x) $\wedge$ House(y) $\wedge$ col2(y) $\wedge$ infrontof(x,y))\\
They played with a col1 ball and col2 bat.& $\exists x \exists y $(Ball(x) $\wedge$ col1(x) $\wedge$ Bat(y) $\wedge$ col2(y) $\wedge$ play-with(they,x) $\wedge$ play-with(they,y))\\
The ball that they played with was col1 and the bat was col2.& $\exists x \exists y $(Ball(x) $\wedge$ col1(x) $\wedge$ Bat(y) $\wedge$ col2(y) $\wedge$ play-with(they,x) $\wedge$ play-with(they,y))\\
col2 was not the color of the ball but of the bat.& $\exists x \exists y $(Ball(x) $\wedge$ $\neg$ col2(x) $\wedge$ Bat(y) $\wedge$ col2(y))\\
col1 was the color of the ball that was hit by the col2 bat.& $\exists x \exists y $(Ball(x) $\wedge$ col1(x) $\wedge$ Bat(y) $\wedge$ col2(y) $\wedge$ was-hit-by(x,y))\\
The ball that was hit by the col2 bat was col1.& $\exists x \exists y $(Ball(x) $\wedge$ col1(x) $\wedge$ Bat(y) $\wedge$ col2(y) $\wedge$ was-hit-by(x,y))\\
The man was wearing a col1 shirt and a col2 jacket.& $\exists x \exists y \exists z$ (Shirt(x) $\wedge$ col1(x) $\wedge$ Jacket(y) $\wedge$ col2(y) $\wedge$ Man(z) $\wedge$ wear(z,x) $\wedge$ wear(z,y))\\
The shirt that the man wore was col1 and the jacket was col2.& $\exists x \exists y \exists z$ (Shirt(x) $\wedge$ col1(x) $\wedge$ Jacket(y) $\wedge$ col2(y) $\wedge$ Man(z) $\wedge$ wear(z,x) $\wedge$ wear(z,y))\\
col2 was not the color of the shirt but of the jacket.& $\exists x \exists y $(Shirt(x) $\wedge$ $\neg$ col2(x) $\wedge$ Jacket(y) $\wedge$ col2(y))\\
col1 was the color of the shirt with the col2 jacket.& $\exists x \exists y$ (Shirt(x) $\wedge$ col1(x) $\wedge$ Jacket(y) $\wedge$ col2(y))\\
The shirt that went with col2 jacket was col1.&$\exists x \exists y$ (Shirt(x) $\wedge$ col1(x) $\wedge$ Jacket(y) $\wedge$ col2(y) $\wedge$ went-with(x,y))\\
The house had a col1 window and a col2 door.&  $ \exists x \exists y \exists z$ (Window(x) $\wedge$ col1(x) $\wedge$ Door(y) $\wedge$ col2(y) $\wedge$ House(z) $\wedge$ had(z,x) $\wedge$ had(z, y))\\
The window that was col1 was next to the door that was col2.& $\exists x \exists y$ (Window(x) $\wedge$ col1(x) $\wedge$ Door(y) $\wedge$ col2(y) $\wedge$ next-to(x,y))\\
col2 was not the color of the window but of the door.& $\exists x \exists y$ (Window(x) $\wedge$ $\neg$ col2(x) $\wedge$ Door(y) $\wedge$ col2(y))\\
col1 was the color of the window next to the col2 door.& $\exists x \exists y$ (Window(x) $\wedge$ col1(x) $\wedge$ Door(y) $\wedge$ col2(y) $\wedge$ next-to(x,y))\\
The window that was next to the col2 door was col1.&  $\exists x \exists y$ (Window(x) $\wedge$ col1(x) $\wedge$ Door(y) $\wedge$ col2(y) $\wedge$ next-to(x,y))\\
A col1 glass was placed on a col2 table.& $\exists x \exists y$ (Glass(x) $\wedge$ col1(x) $\wedge$ Table(y) $\wedge$ col2(y) $\wedge$ placed-on(x,y))\\
The glass that was col1 was placed on a table that was col2.&$\exists x \exists y$ (Glass(x) $\wedge$ col1(x) $\wedge$ Table(y) $\wedge$ col2(y) $\wedge$ placed-on(x,y))\\
col2 was not the color of the glass but of the table.& $\exists x \exists y$ (Glass(x) $\wedge$ $\neg$ col2(x) $\wedge$ Table(y) $\wedge$ col1(y))\\
col1 was the color of the glass placed on the col2 table.&  $\exists x \exists y$ (Glass(x) $\wedge$ col1(x) $\wedge$ Table(y) $\wedge$ col2(y) $\wedge$ placed-on(x,y))\\
The glass that was placed on a col2 table was col1.& $\exists x \exists y$ (Glass(x) $\wedge$ col1(x) $\wedge$ Table(y) $\wedge$ col2(y) $\wedge$ placed-on(x,y))\\ \hline
\end{tabular}
\caption{Templates and FOL for $D_{1,1}$ dataset. \emph{col1} and \emph{col2} refer to two distinct colors.}
\label{tab:d1-1}
\end{table*}

\newpage
\twocolumn

\begin{table}[h]
\begin{tabular}{@{}ll@{}}
\toprule
\multicolumn{2}{c}{GPUs}             \\ \midrule
\multicolumn{2}{c}{4 NVIDIA Volta V100} \\ \midrule\midrule
\multicolumn{2}{c}{Hyperparameters} \\ \midrule
Training epochs             & 10     \\
batch size                  & 4     \\
optimizer                   & Adam     \\
learning rate               & 2e-4   \\
\multirow{2}{*}{learning rate scheduler}     & linear warm-up and     \\
                                            & cosine annealing \\
warm-up ratio               & 0.03    \\
gradient clipping           &  0.3    \\
lora r                      & 64     \\
lora (alpha)                & 16     \\
lora dropout ratio          & 0.1     \\
\multirow{2}{*}{lora target modules}         &  Only Attention Blocks\\ 
                                        & (q\_proj, v\_proj)    \\
quantization    & 4-bit NormalFloat \\ \bottomrule
\end{tabular}
\caption{\label{tab:model-details}Details on computing resources and hyperparameters for finetuning LLMs for FOL translation.}
\end{table}

Table~\ref{tab:model-details} gives the hyperparameters used for finetuning LLMs for FOL translation along with the computing resources. We adapt the finetuning code from the following repository\footnote{\url{https://github.com/mlabonne/llm-course/blob/main/Fine_tune_Llama_2_in_Google_Colab.ipynb}}. Finetuning LLMs as per the hyperparameters given in Table~\ref{tab:model-details} took around 5 hours.

\end{document}